\documentclass[12pt, a4paper]{article}
\usepackage[utf8]{inputenc}

\usepackage{subcaption}
\usepackage{float}
\usepackage{balance}
\usepackage{graphicx}
\usepackage{multicol}
\usepackage{array}
\usepackage{comment}
\usepackage[english]{babel}

\usepackage{amsthm}
\usepackage{amssymb,amsopn,amsmath,amstext,amsfonts,algorithmic,psfrag}
\usepackage[T1]{fontenc}
\usepackage{mwe}    
\usepackage{subfig} 
\usepackage{cite}   
\usepackage[legalpaper, margin=1.5cm]{geometry}

\title{\LARGE \bf
NeuroHJR: Hamilton-Jacobi Reachability-based Obstacle Avoidance in Complex Environments with Physics-Informed Neural Networks
}

\author{Granthik Halder$^{1\#}$, Rudrashis Majumder$^{2*\#}$, Rakshith M R$^{3}$, \\ Rahi Shah$^{4}$ and Suresh Sundaram$^{3}$
    \thanks{$^{*}$Corresponding Author}
    \thanks{$^{\#}$Equal Contribution}    
	\thanks{$^{1}$Granthik Halder from the Department of Physical Science, Indian Institute of Science Education and Research, Kolkata, India, has worked as a Research Intern with the Department of Aerospace Engineering, Indian Institute of Science, Bangalore, India.
		{\tt\small gh23ms065@iiserkol.ac.in} } 
	\thanks{$^{2}$Rudrashis Majumder is with the Department of Computer Science and Engineering, Shiv Nadar University, Chennai, India 
		{\tt\small rudrashism@snuchennai.edu.in}}%
        	\thanks{$^{3}$M R Rakshith and Suresh Sundaram are with the Department of Aerospace Engineering, Indian Institute of Science, Bangalore 560012,
		Karnataka, India 
		{\tt\small mrrakshith.iisc@gmail.com, vssuresh@iisc.ac.in}}%
        	\thanks{$^{4}$Rahi Shah from the Department of Computer Science Engineering, Ahmedabad University, India, has worked as a Research Intern with the Department of Aerospace Engineering, Indian Institute of Science, Bangalore, India.
		{\tt\small rahi.s@ahduni.edu.in}. }%
}

\begin{document}
\maketitle
\pagenumbering{arabic}
\begin{abstract}
Autonomous ground vehicles (AGVs) must navigate safely in cluttered environments while accounting for complex dynamics and environmental uncertainty. Hamilton-Jacobi Reachability (HJR) offers formal safety guarantees through the computation of forward and backward reachable sets, but its application is hindered by poor scalability in environments with numerous obstacles. In this paper, we present a novel framework called NeuroHJR that leverages Physics-Informed Neural Networks (PINNs) to approximate the HJR solution for real-time obstacle avoidance. By embedding system dynamics and safety constraints directly into the neural network loss function, our method bypasses the need for grid-based discretization and enables efficient estimation of reachable sets in continuous state spaces. We demonstrate the effectiveness of our approach through simulation results in densely cluttered scenarios, showing that it achieves safety performance comparable to that of classical HJR solvers while significantly reducing the computational cost. This work provides a new step toward real-time, scalable deployment of reachability-based obstacle avoidance in robotics.

\end{abstract}

\section{Introduction}\label{sec:intro}
Autonomous ground vehicles (AGVs) are increasingly being deployed in various domains \cite{hebert2012intelligent} such as logistics, agriculture, and surveillance, where safe and efficient navigation is critical. Crucial to their autonomy is the ability to plan the vehicle's motion strategically while responding to the surrounding environment. In particular, real-time obstacle avoidance has emerged as a key factor facilitating reliable and safe autonomous navigation \cite{moon2010study}. However, this task requires perception of the environment and timely generation of control actions that respect vehicle dynamics and safety constraints.

Unmanned ground vehicles (UGVs), in both civilian and military applications, face significant challenges in navigating complex terrains with static and dynamic obstacles. Various obstacle avoidance algorithms have been developed, including Artificial Potential Fields (APF) \cite{sun2017collision}, collision cone methods \cite{xu2020dynamic}, and Model Predictive Control (MPC) \cite{park2009obstacle}. More recently, Control Barrier Functions (CBFs) have gained popularity for their ease of implementation, theoretical safety guarantees, and compatibility with feedback-based control \cite{ames2016control, ames2019control}. However, each of these methods has inherent limitations: APF suffers from local minima and instability near dynamic obstacles, collision cone approaches are computationally intensive in densely cluttered environments, and MPC, despite offering optimality, is often too slow for real-time deployment. Although CBFs allow for real-time operation with safety guarantees, they typically offer only myopic safety guarantees and are sensitive to parameter tuning.

In contrast, Hamilton-Jacobi Reachability (HJR) \cite{tomlin2003computational, herbert2020safe} offers a rigorous framework for safety-critical planning by computing forward and backward reachable sets that capture all future states leading to unsafe conditions under worst-case scenarios. This makes HJR particularly attractive for obstacle avoidance in densely cluttered environments. Unlike the above-mentioned existing methods, HJR provides global safety guarantees by integrating system dynamics, control constraints, and obstacle geometry into a unified formulation. However, the practical use of HJR is hindered by its poor scalability: its computational cost increases exponentially with the number of obstacles, as each obstacle must be incorporated into the target set representation. This renders traditional HJR solvers infeasible for real-time applications in complex environments.

In this work, we propose a Physics-Informed Neural Network (PINN)-based approach \cite{raissi2019physics, majumder2024oa, majumder2024safe}, called NeuroHJR, to approximate the Hamilton-Jacobi Reachability solution for obstacle avoidance in cluttered settings. In a cluttered environment with dense obstacles, Hamilton-Jacobi Reachability (HJR) becomes computationally intensive because each obstacle must be encoded into the level set representation, significantly increasing the complexity of the value function and its evolution. This leads to higher memory usage and longer computation times, especially when using grid-based solvers. Physics-Informed Neural Networks (PINNs) address this challenge by representing the value function continuously without explicit gridding, allowing the model to generalize over complex obstacle configurations. As a result, PINNs enable scalable and efficient approximation of HJR solutions, making real-time obstacle avoidance feasible even in densely cluttered environments. The proposed approach bridges the gap between the formal rigor of reachability analysis and the scalability of learning-based methods. This work demonstrates the effectiveness of the proposed framework in simulated scenarios, showing its potential to maintain real-time safety while significantly reducing computational cost. The main contributions of this work are as follows:
\begin{itemize}
    \item A novel PINN-based framework is proposed to approximate the Hamilton-Jacobi Reachability (HJR) solution for real-time obstacle avoidance in cluttered environments.

    \item The method embeds system dynamics and safety constraints in the training loss, enabling scalable reachability computation without grid-based discretization.

    \item The proposed approach handles multiple obstacles efficiently, overcoming the scalability limitations of classical HJR solvers.

    \item We demonstrate through simulations that our method achieves safety performance comparable to traditional HJR, while providing much better real-time evaluation.

    \item An ablation study is also presented to evaluate the impact of varying sensor radius on the NeuroHJR performance.

\end{itemize}
The remainder of this paper is organized as follows. Section~\ref{sec:preli_ps} introduces the necessary mathematical preliminaries and outlines the problem formulation. The proposed methodology, NeuroHJR, is detailed in Section~\ref{sec:NeuroHJR}. Section~\ref{sec:mainresult} presents simulation results and an ablation study to evaluate the performance of NeuroHJR. Finally, Section~\ref{sec:conclusions} summarizes the key findings and discusses directions for future work.

\section{Preliminaries and Problem Statement}\label{sec:preli_ps}
This section introduces the preliminary concepts on Hamilton-Jacobi Reachability (HJR) Physics-Informed Neural Networks (PINN). Then, the problem setting addressed in this work is presented, including the system model, obstacle assumptions, and the safety specification to be approximated using a PINN.

\subsection{Hamilton-Jacobi Reachability}
Hamilton-Jacobi Reachability (HJR) \cite{tomlin2003computational, herbert2020safe} is a powerful framework for analyzing safety and reachability in dynamical systems. It provides mathematical tools to compute safe sets and optimal control strategies and is particularly effective for obstacle avoidance and safety-critical control problems.

The HJR framework is governed by the Hamilton-Jacobi-Bellman (HJB) equation, which describes the evolution of the value function $V(x,t)$:

\begin{align}
\frac{\partial V}{\partial t} + H\left(x, \frac{\partial V}{\partial x}\right) &= 0
\label{eq:HJBeq}
\end{align}
where $V(x,T) = g(x)$, for some final time $T$ and the Hamiltonian is defined as
\begin{align}
H\left(x, \frac{\partial V}{\partial x}\right) = \min_{u \in \mathcal{U}} \left\{ \frac{\partial V}{\partial x} \cdot f(x,u) \right\}
\end{align}

The key components include $V(x,t)$ as the value function representing safety/cost, $g(x)$ as the terminal condition, and $H(\cdot)$ as the Hamiltonian function.

\subsubsection{Forward Reachability Set}
The forward reachable set from the initial set $\mathcal{S}_0$ at time $t$ is defined as:

\begin{align}
\mathcal{R}_F(t) = \{x(t) : x(0) \in \mathcal{S}_0, \exists u(\cdot)\}
\end{align}

This represents the set of all states reachable from $\mathcal{S}_0$ within time $t$, where control tries to maximize reachability and disturbance tries to minimize reachability. 

The HJB formulation for forward reachability is as follows:
\begin{align}
\frac{\partial V_F}{\partial t} + \min_{u \in \mathcal{U}} \left\{ \frac{\partial V_F}{\partial x} \cdot f(x,u) \right\} &= 0
\end{align}
\begin{align}
\textnormal{where}, \quad V_F(x,0) &= \min\{\|x - y\| : y \in \mathcal{S}_0\}
\end{align}

\subsubsection{Backward Reachability Set}
The backward reachable set for the target set $\mathcal{T}$ at time $t$ is defined as:

\begin{align}
\mathcal{R}_B(t) = \{x(0) : x(T) \in \mathcal{T}, \exists u(\cdot)\}
\end{align}

This represents the set of all initial states that can reach $\mathcal{T}$ within time $T-t$, where the control tries to ensure that the target is reached, and the disturbance tries to prevent the target from reaching.

The HJB formulation for backward reachability is:
\begin{align}
\frac{\partial V_B}{\partial t} + \min_{u \in \mathcal{U}} \left\{ \frac{\partial V_B}{\partial x} \cdot f(x,u) \right\} &= 0
\end{align}
\begin{align}
\textnormal{where}, \quad V_B(x,T) &= \min\{\|x - y\| : y \in \mathcal{T}\}
\end{align}

For obstacle avoidance, both approaches are combined using a composite value function:
\begin{align}
V(x,t) = \max\{V_F(x,t), -V_B(x,t)\}
\label{eq:composite_V}
\end{align}
A trajectory is safe if $V(x(t),t) \geq 0$ for all $t \in [0, T]$.

\subsection{Physics-Informed Neural Networks}

Physics-Informed Neural Networks (PINNs) \cite{raissi2019physics} integrate neural networks with physical laws, typically in the form of partial differential equations (PDEs), to constrain learning and produce solutions that satisfy the governing equations of the system \cite{raissi2019physics}. Unlike purely data-driven models, PINNs incorporate prior knowledge of system dynamics, enabling accurate and physically consistent predictions even in regions with sparse or noisy data. Let a physical system be described by a PDE of the form
\begin{equation}
    \mathcal{N}[u(\mathbf{x}, t)] = 0,\quad \mathbf{x} \in \Omega, \; t \in [0, T],
\end{equation}
where \( \mathcal{N}[\cdot] \) is a nonlinear differential operator, and \( u(\mathbf{x}, t) \) is the unknown solution. A PINN approximates \( u(\mathbf{x}, t) \) using a neural network \( u_\theta(\mathbf{x}, t) \) with trainable parameters \( \theta \), and incorporates the PDE residual directly into the loss function \cite{majumder2024oa, majumder2024safe}. The total loss typically takes the form of
\begin{equation}
    \mathcal{L}(\theta) = \lambda_{data}\mathcal{L}_{\text{data}} + \lambda_{physics}\mathcal{L}_{\text{PDE}},
\end{equation}
where \( \mathcal{L}_{\text{data}} \) enforces agreement with available measurements, and 
\[
    \mathcal{L}_{\text{PDE}} = \left\| \mathcal{N}[u_\theta(\mathbf{x}, t)] \right\|^2
\]
penalizes deviations from the governing physics. This physics-informed training paradigm enables the network to infer solutions that are consistent with both observed data and the underlying physical principles, making PINNs particularly effective for solving forward and inverse problems in scenarios with limited or noisy data.

In our work, we utilize the PINN to solve the time-invariant (steady-state) form of the HJB equation for obstacle avoidance in cluttered environments, where the network simultaneously learns the optimal control policy and the value function for safe navigation. 

\subsection{Problem Definition}
A single UGV is considered to study the efficacy of the proposed NeuroHJR approach for an obstacle avoidance problem. The area for the UGV movement is considered with specific start and goal locations, and several cylindrical obstacles. The vehicle is expected to successfully reach the goal location without colliding with any of the obstacles. The UGV is assumed to reach the goal once its distance from the goal is less than a specific threshold value.

To explain the idea mathematically, let 
\begin{equation}
    d = || p_{ugv}(t) - p_{obs}(t)|| = \sqrt{(x - x_{o})^2 + (y - y_{o})^2}
\end{equation}
be the distance between the UGV and any obstacle located at coordinate $(x_{o}, y_{o})$. For safe path planning of UGVs, $d > d_{threshold}$ is required, be it a static or a dynamic obstacle in the environment. However, avoiding obstacles is not the only concern of a successful mission. After starting the journey, the UGV should reach the goal in a finite amount of time. Therefore, it is expected that
\begin{equation}
    || p_{ugv}(t) - p_{goal}(t)|| \rightarrow 0
\end{equation}
occurs at some finite time $t = T$. 

\subsubsection{System Model}
We consider velocity-controlled UGV for obstacle avoidance represented by a holonomic first-order model as
\begin{equation}
    \dot{\textbf{x}}(t) = \textbf{u}(t)
\label{eq:model}
\end{equation}
where $\textbf{x} = [P_x, P_y]$ represents the position of the UGV on the ground and $\textbf{u} = [u_x, u_y]$ is the respective velocity that acts as a control input. Being a ground vehicle, its motion is considered in a two-dimensional environment.

\section{NeuroHJR Methodology}\label{sec:NeuroHJR}
In this section, we present the formulation of NeuroHJR, our proposed PINN-based framework for approximating Hamilton-Jacobi Reachability (HJR) solutions in the context of obstacle avoidance. The approach embeds the system dynamics and safety constraints into the loss function of a neural network, enabling continuous and scalable approximation of the value function without relying on discretized grids. We begin by formulating the PDE for the HJR problem, followed by the design of the PINN architecture and training strategy.

\subsection{PDE for Hamilton Jacobi Reachability}
The value function $V(x,t)$ is the cornerstone of Hamilton-Jacobi reachability and serves as the mathematical foundation for our approach. The value function represents:
\begin{equation}
V(x,t) = \min_{u(\cdot)} J(x,t,u)
\end{equation}
where $J$ represents the cost-to-go from state $x$ at time $t$. We employ a composite value function constructed from both the forward and backward reachable sets, as defined in Equation \eqref{eq:composite_V}.

In the context of our obstacle avoidance problem, the state variables 
$(x,y)$ in the HJB equation correspond to the UGV's position coordinates $(P_x,P_y)$. For optimal control problems, the value function satisfies the Hamilton-Jacobi-Bellman (HJB) equation:
\begin{equation}
\frac{\partial V}{\partial t} + \min_{u \in \mathcal{U}} \left\{ \frac{\partial V}{\partial P_x} f_x(P_x, P_y, u) + \frac{\partial V}{\partial P_y} f_y(P_x, P_y, u) \right\} = 0
\end{equation}
which represents an expanded form of the general HJB equation presented in Equation \eqref{eq:HJBeq}.

For our system with robot dynamics $f_x = u_x$ and $f_y = u_y$ as defined in Equation \eqref{eq:model}, the HJB equation becomes:
\begin{equation}
\frac{\partial V}{\partial t} + \min_{u_x, u_y} \left\{ \frac{\partial V}{\partial P_x} u_x + \frac{\partial V}{\partial P_y} u_y \right\} = 0
\end{equation}

In our case, since the starting position, goal position, and obstacle configuration remain static, the problem is time-invariant. Therefore, we consider the steady-state case where $\frac{\partial V}{\partial t} = 0$:
\begin{equation}
\min_{u_x, u_y} \left\{ \frac{\partial V}{\partial P_x} u_x + \frac{\partial V}{\partial P_y} u_y \right\} = 0
\end{equation}
We now proceed to solve this PDE using NeuroHJR.

\begin{figure}[htbp]
\centering
\includegraphics[width=0.45\textwidth]{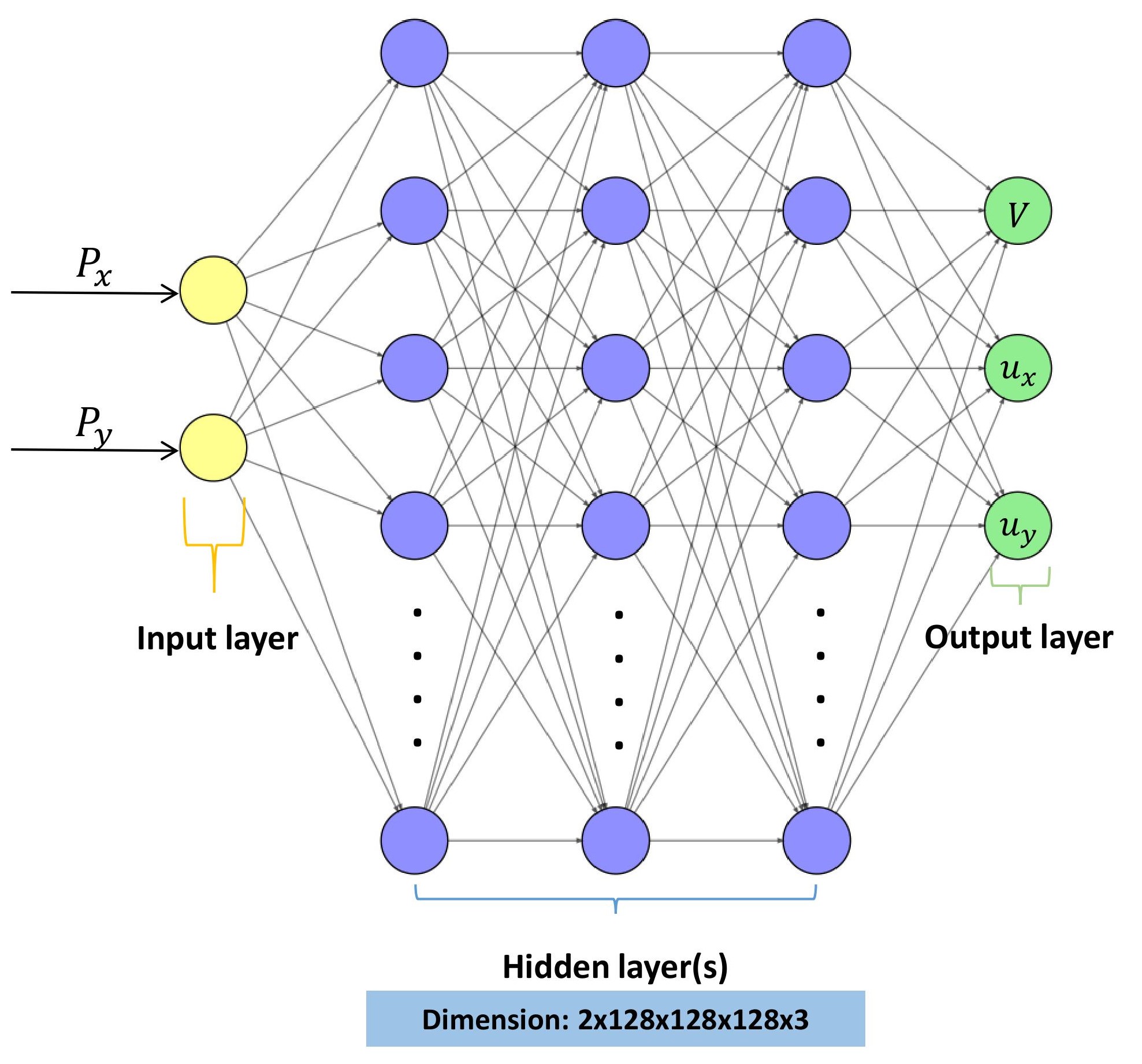}
\caption{NeuroHJR Architecture}
\label{fig:NN-Structure}
\end{figure}

\subsection{NeuroHJR Architecture}

In NeuroHJR, the Physics-Informed Neural Network (PINN) consists of a carefully designed network structure optimized to solve the Hamilton-Jacobi-Bellman equation, as shown in Fig. \ref{fig:NN-Structure}. The network takes a 2D input $(P_x, P_y) \in \mathbb{R}^2$, representing position coordinates at each time step $\Delta t = 0.1$ second and processes it through 3 hidden layers with 128 neurons each, using $\tanh$ activation functions. The architecture then branches into two output heads: a control head with $\tanh$ activation producing bounded outputs $(u_x, u_y)$, representing the control inputs (velocities) bounded by $|u_x|, |u_y| \leq 1$, and a value head with linear activation producing an unbounded scalar output $V \in \mathbb{R}$.

The mathematical representation of the network is:
$$\text{NeuroHJR}: (P_x, P_y) \mapsto (u_x, u_y, V)$$
where the network learns both the optimal control policy and value function simultaneously. NeuroHJR uses automatic differentiation to compute the value function gradients:
\begin{equation}
\frac{\partial V}{\partial P_x} = \frac{\partial V_{\text{NN}}(P_x, P_y; \theta)}{\partial P_x}, \quad \frac{\partial V}{\partial P_y} = \frac{\partial V_{\text{NN}}(P_x, P_y; \theta)}{\partial P_y}
\label{eq:differentiation}
\end{equation}
where $\theta$ represents the neural network parameters.

\subsection{Training and Loss Function}
The training process of NeuroHJR involves minimizing a loss function that incorporates governing physical laws and observational data. This is achieved by using collocation points, which are strategically chosen locations within the domain where the PDE residual is enforced to be approximately zero, ensuring that the solution satisfies the equation at those points. The training data collocation points are generated by sampling $N_c = 8000$ random points from the safe region $\mathcal{S}$, along with $N_b = 5000$ points concentrated near obstacle boundaries to improve boundary resolution. This approach offers two key advantages: (1) it naturally handles complex obstacle shapes due to its flexible geometric formulation, and (2) it achieves computational efficiency through parallel evaluation at all spatial points. The model is trained using the Adam optimizer with a learning rate of $\alpha = 0.001$, for a total of $E = 10000$ epochs.

The total loss function is composed of four components:
\begin{equation}
\mathcal{L}_{\text{total}} = \mathcal{L}_{\text{PDE}} + \mathcal{L}_{\text{value}} + \mathcal{L}_{\text{obstacle}} + \mathcal{L}_{\text{goal}}
\end{equation}
Each of the components is defined as follows.
\subsubsection{PDE Residual Loss}
This is the most critical component that enforces the HJB equation:
\begin{equation}
\mathcal{L}_{\text{PDE}} = \frac{1}{N} \sum_{i=1}^{N} \left( \frac{\partial V}{\partial P_x} u_x^{(i)} + \frac{\partial V}{\partial P_y} u_y^{(i)} \right)^2
\end{equation}
where $(P_x^{(i)}, P_y^{(i)})$ are the $i$-th collocation points, $u_x^{(i)}, u_y^{(i)}$ are PINN-predicted controls, and $\frac{\partial V}{\partial P_x}, \frac{\partial V}{\partial P_y}$ are given by Equation \eqref{eq:differentiation}. This loss ensures that the learned control policy minimizes the Hamiltonian, effectively solving the HJB PDE at each collocation point.

\subsubsection{Value Function Loss}
\begin{equation}
\mathcal{L}_{\text{value}} = \frac{1}{N} \sum_{i=1}^{N} \left( V_{\text{pred}}^{(i)} - V_{\text{true}}^{(i)} \right)^2
\end{equation}
where the $V_{\text{pred}}$ is the value function predicted by the neural network and $V_{\text{true}}$ is the value function computed using forward and backward reachability sets, given by Equation \eqref{eq:composite_V}. This loss ensures the neural network accurately predicts the value function by minimizing the mean squared error between predicted and ground truth values.

\subsubsection{Obstacle Avoidance Loss}
\begin{equation}
\mathcal{L}_{\text{obstacle}} = \frac{1}{N} \sum_{i=1}^{N} \left( \max(0, R_d - d_i) \right)^2
\end{equation}
where $d_i = \sqrt{(P_x^{(i)} - x_o)^2 + (P_y^{(i)} - y_o)^2}$ is the distance from point $i$ to the obstacle center and $R_d$ represents the radius of the entire unsafe region, which is the original obstacle radius $R$ plus a safety margin $\delta$. This loss penalizes the network when it predicts trajectories that come too close to obstacles by applying a penalty proportional to the square of the distance violation within the radius of the unsafe zone.

\subsubsection{Goal Convergence Loss}
\begin{equation}
\mathcal{L}_{\text{goal}} = -\frac{1}{N} \sum_{i=1}^{N} \left\langle \vec{u}^{(i)}, \vec{r}_{\text{goal}}^{(i)} \right\rangle
\end{equation}
This loss encourages alignment between the control vector $\vec{u}^{(i)} = (u_x^{(i)}, u_y^{(i)})$ and the unit direction vector toward the goal $\vec{r}^{(i)}_{\text{goal}} = \frac{\langle x_g -P_ x^{(i)}, y_g - P_y^{(i)} \rangle}{\|(x_g - P_x^{(i)}, y_g - P_y^{(i)})\|}$, where $(x_g, y_g)$ is the goal position and $\|x_g - P_x^{(i)}, y_g - P_y^{(i)}\| = \sqrt{(x_g - P_x^{(i)})^2 + (y_g - P_y^{(i)})^2}$, by maximizing their dot product $\langle \vec{u}, \vec{r}_{\text{goal}} \rangle > 0$. The negative sign ensures that maximizing alignment corresponds to minimizing the loss.

The vehicle initially follows a straight-line path to the goal using linear motion. Upon sensing an obstacle, it switches to NeuroHJR-generated control to avoid collision, then resumes the shortest path toward the goal. This behavior is illustrated in the next section.

\begin{figure*}[h]
\centering
\begin{subfigure}{0.35\textwidth}  
\includegraphics[width=\textwidth]{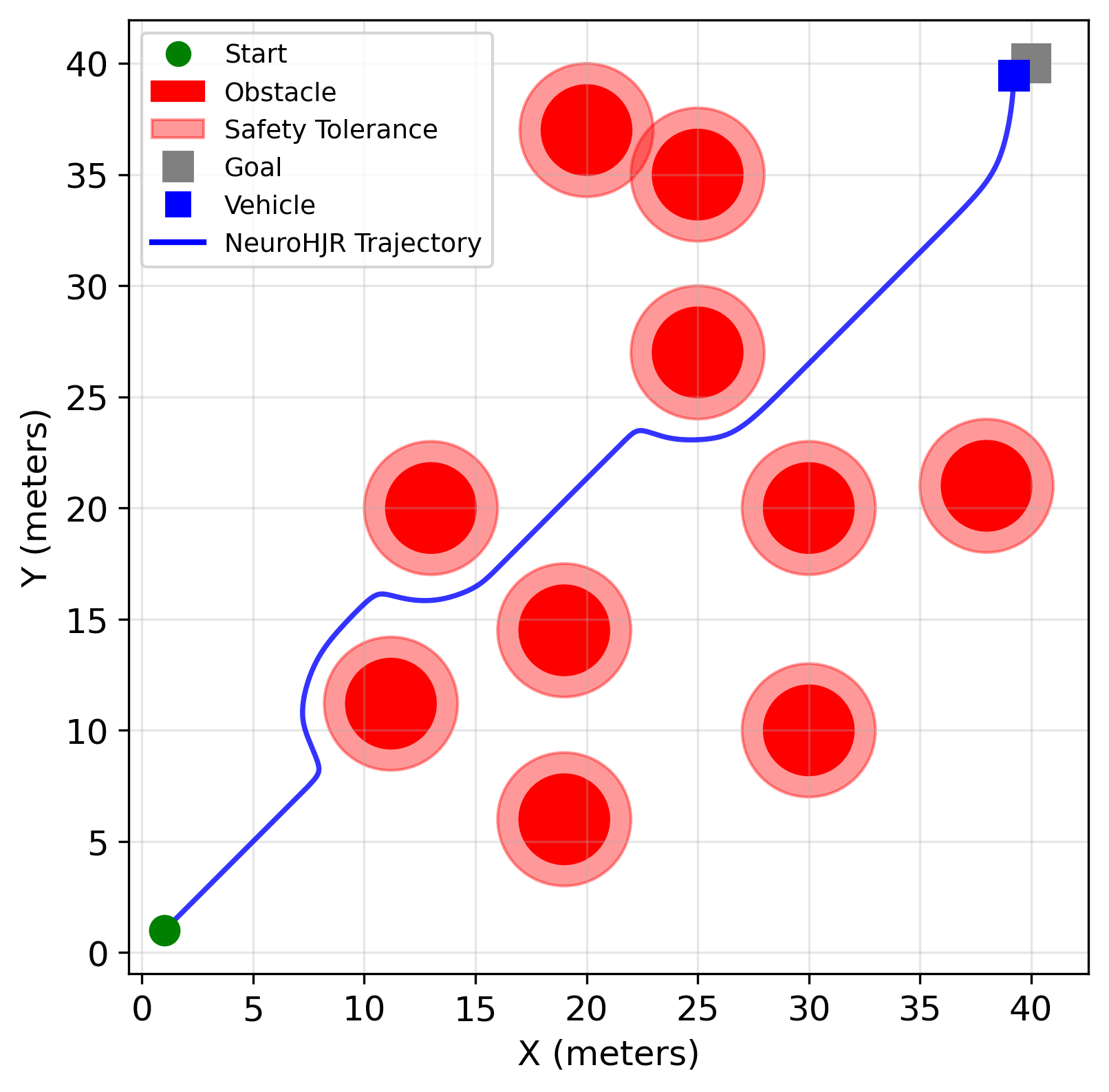}
\caption{Cluttered Environment 1}
\label{fig:sim1}
\end{subfigure}
\begin{subfigure}{0.35\textwidth}  
\includegraphics[width=\textwidth]{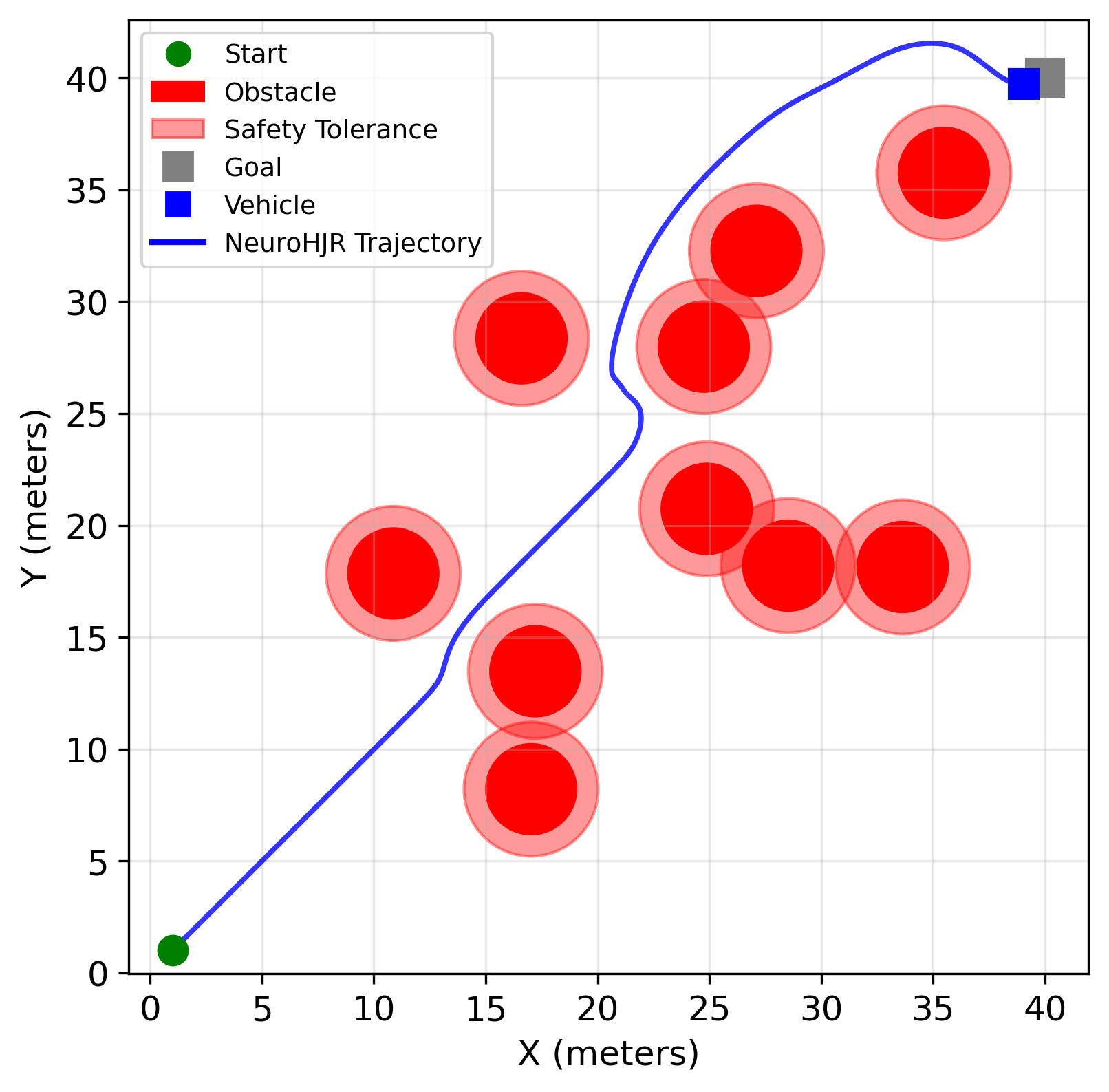}
\caption{Cluttered Environment 2}
\label{fig:sim2}
\end{subfigure}
\caption{Obstacle avoidance using NeuroHJR}
\label{fig:simulation_results}
\end{figure*}

\begin{figure*}[h]
\centering
\begin{subfigure}{0.32\textwidth}
\includegraphics[width=\textwidth]{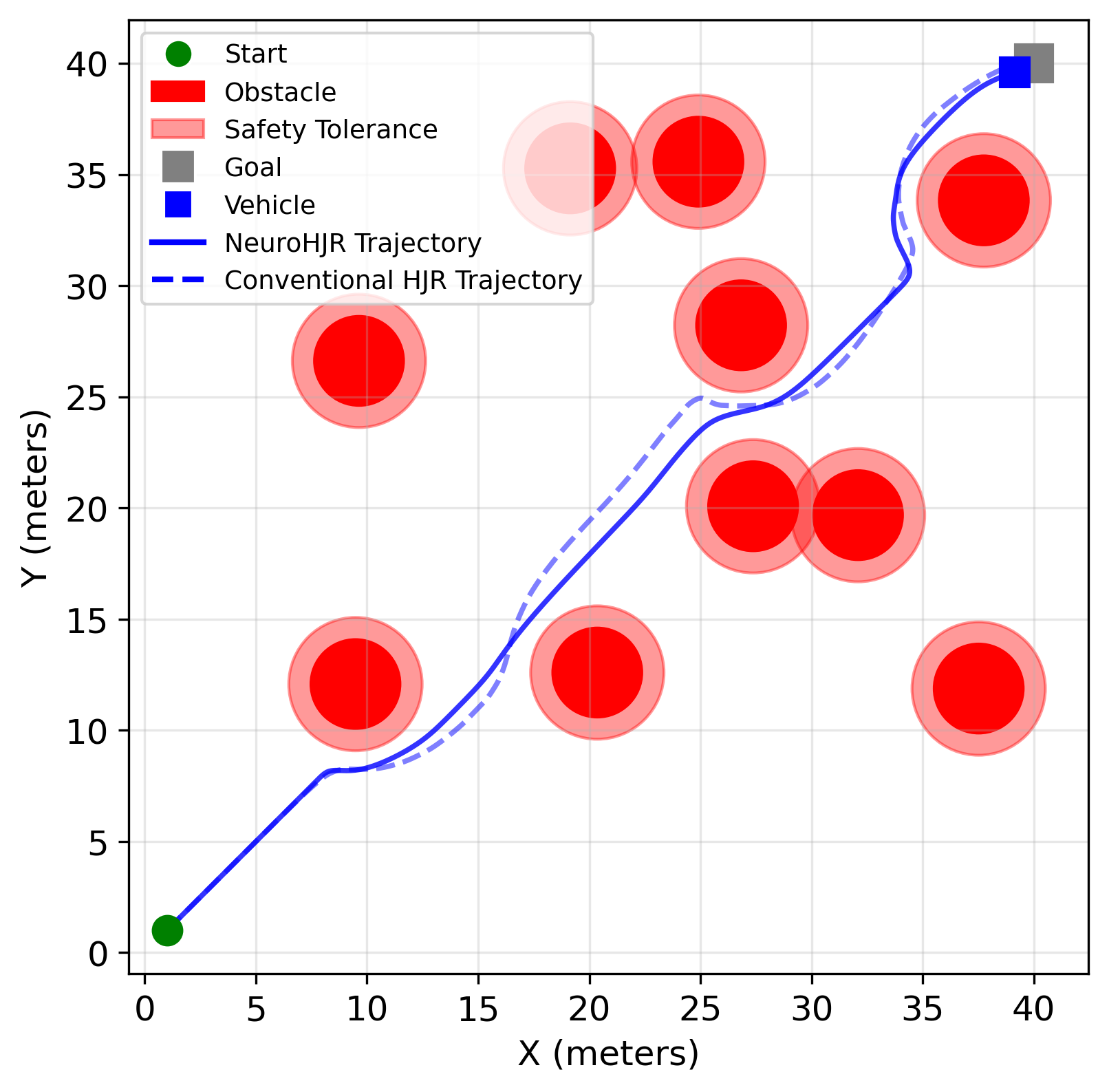}
\caption{Scenario 1}
\label{fig:comp1}
\end{subfigure}
\hfill
\begin{subfigure}{0.32\textwidth}
\includegraphics[width=\textwidth]{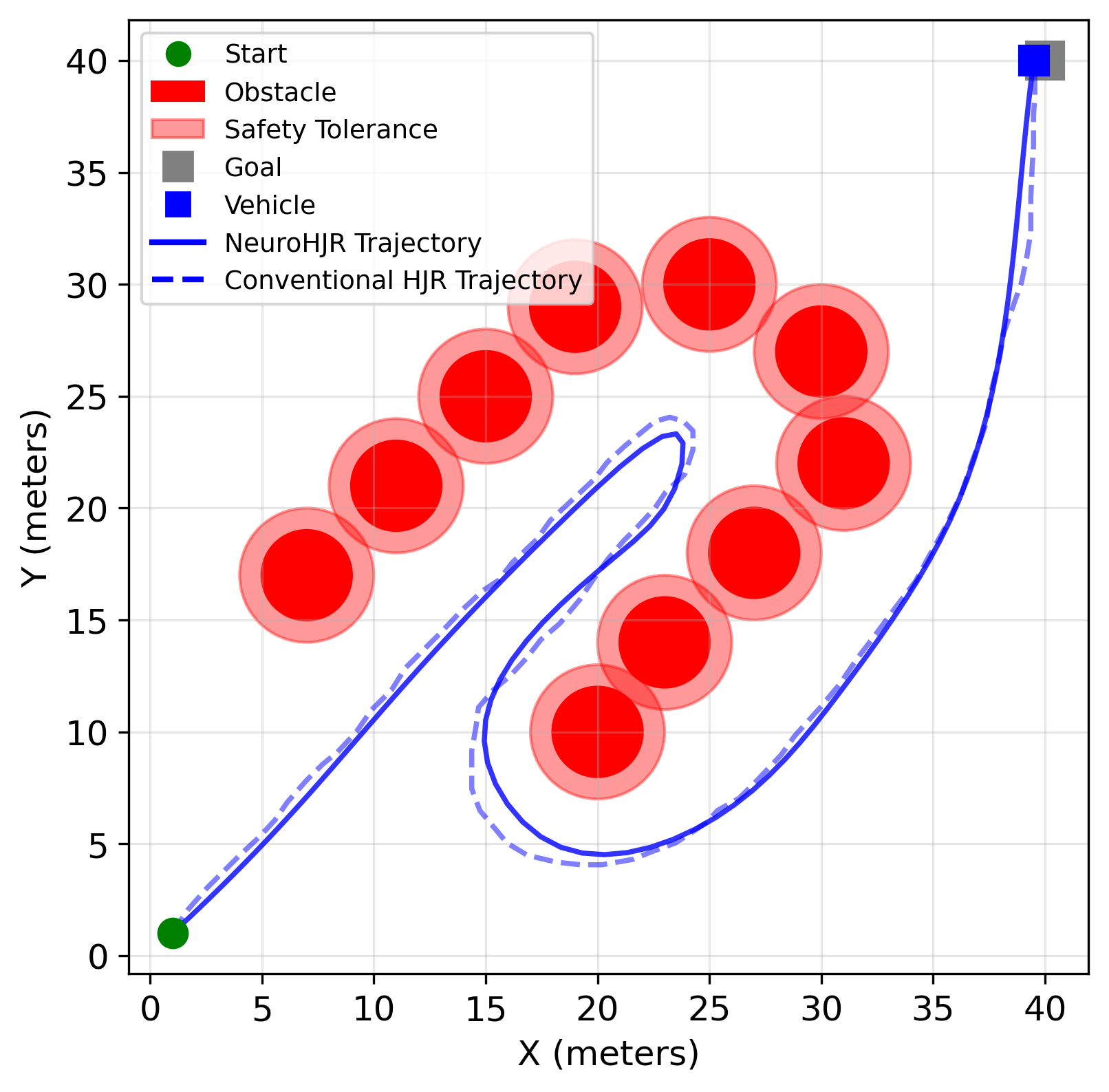}
\caption{Scenario 2}
\label{fig:comp2}
\end{subfigure}
\hfill
\begin{subfigure}{0.32\textwidth}
\includegraphics[width=\textwidth]{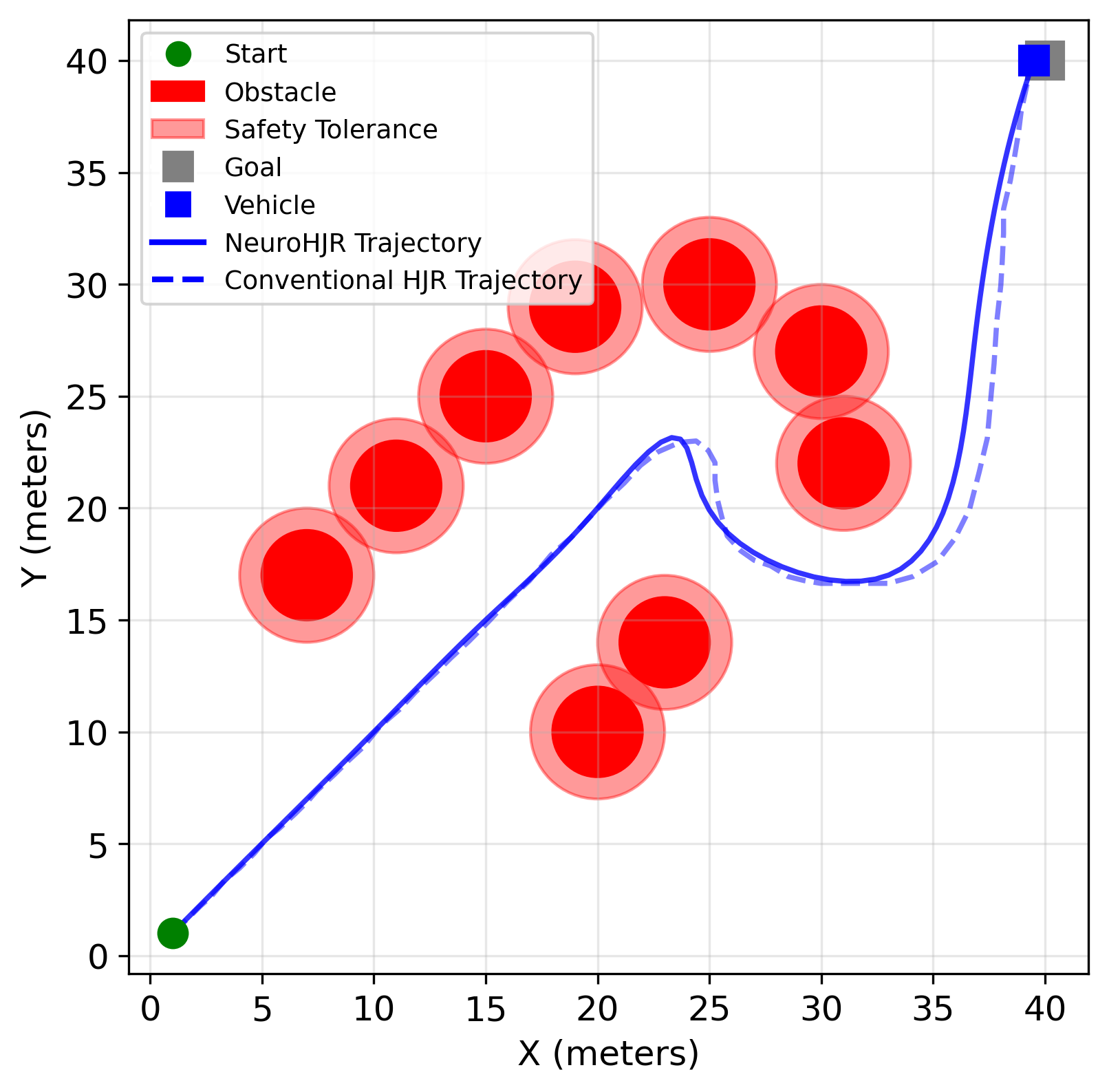}
\caption{Scenario 3}
\label{fig:comp3}
\end{subfigure}
\caption{Trajectory Comparison: NeuroHJR vs Conventional HJR}
\label{fig:comparison_results}
\end{figure*}
\section{Performance Evaluation}\label{sec:mainresult}

In this section, we provide the results for the evaluation of NeuroHJR using Python simulations.

\subsection{Numerical Simulations}
The proposed NeuroHJR framework was evaluated through comprehensive numerical simulations in cluttered environments. The simulation parameters are configured as follows: the system operates within a 45\textit{m} × 45\textit{m}, the robot starts at position $(x_s, y_s) = (1, 1)$, with the goal located at $(x_g, y_g) = (40, 40)$; ten cylindrical obstacles (circular in 2D from the top view) are randomly placed in the environment, each with a radius of $R = 2.0$ meters; a safety tolerance of $\delta = 1$ meters is applied, defining an unsafe region around each obstacle with an expanded radius $R_d = R + \delta = 3.0$ meters.

The UGV utilizes a sensor range of 5 meters to detect obstacles within its vicinity and triggers the NeuroHJR-based avoidance control when it is on the verge of entering any unsafe region around the obstacle. Fig. \ref{fig:sim1} and \ref{fig:sim2} show the trajectories of the UGV in two different environments from the starting to the goal locations while avoiding obstacles. These simulation results validate that the NeuroHJR framework can effectively solve complex navigation problems in cluttered environments.

\subsection{Baseline Comparison}
A comprehensive performance comparison was conducted between the proposed novel NeuroHJR method and the conventional HJR solver-based approach. The evaluation considers both computational efficiency and path optimization characteristics as the performance metrics. The comparison of the respective trajectories in three different scenarios are shown in Fig. \ref{fig:comp1} - \ref{fig:comp3}.



\begin{table}[h]
\centering
\begin{tabular}{|c|p{2cm}|p{3cm}|p{3cm}|}
\hline
Scenario & NeuroHJR (s) & Conventional HJR (s) & Travel Time Reduction (\%) \\
\hline
Scenario 1 & 44.9 & 59.3 & 24.3 \\
Scenario 2 & 77.3 & 105.4 & 26.7 \\
Scenario 3 & 54.7 & 70.8 & 22.8 \\
\hline
Average & - & - & 24.6 \\
\hline
\end{tabular}
\caption{Computational Speed Comparison}
\label{tab:speed_comparison}
\end{table}

The results in Table~\ref{tab:speed_comparison} for 50 Monte Carlo runs demonstrate that NeuroHJR consistently outperforms the conventional HJR approach, achieving a reduction of 24.6\% in travel time of the UGV across all the tested scenarios, as shown in Fig. \ref{fig:comp1}- \ref{fig:comp3}. The UGV following the conventional HJR approach requires significantly more time to reach the goal compared to NeuroHJR due to the higher computational burden of the conventional approach.

\begin{table}[h]
\centering
\begin{tabular}{|c|p{2cm}|p{3cm}|p{3cm}|}
\hline
Scenario & NeuroHJR (m) & Conventional HJR (m) & Path Efficiency (\%) \\
\hline
Scenario 1 & 57.24 & 59.10 & 3.15 \\
Scenario 2 & 100.48 & 104.24 & 3.61 \\
Scenario 3 & 68.89 & 70.28 & 1.98 \\
\hline
Average & - & - & 2.91 \\
\hline
\end{tabular}
\caption{Path Length Efficiency Comparison}
\label{tab:path_comparison}
\end{table}

As evident from Table~\ref{tab:path_comparison}, the UGV with our proposed NeuroHJR method consistently travels shorter paths compared to conventional approach to reach the goal, as shown in Fig. \ref{fig:comp1}- \ref{fig:comp3}. The average path length reduction is 2.91\% with NeuroHJR. While this improvement may appear modest, it represents significant gains in terms of energy efficiency and mission completion time for robotic applications, as observed in Table~\ref{tab:speed_comparison}.

\subsection{Ablation Studies}
\begin{figure}[htbp]
\centering
\includegraphics[width=0.35\textwidth]{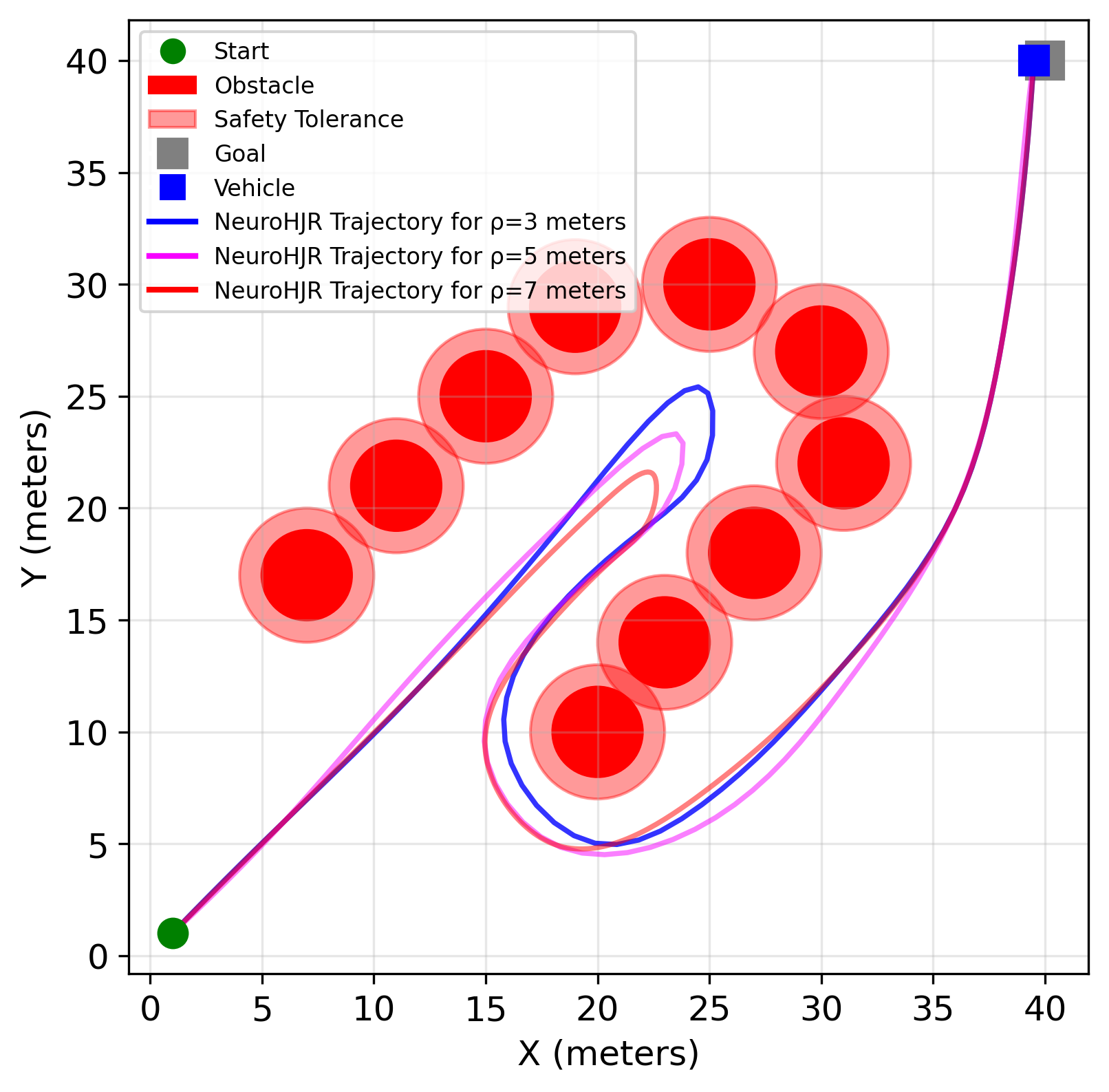}
\caption{Comparison of Trajectories for Different Sensor Radii}
\label{fig:ablation_results}
\end{figure}

To evaluate the impact of sensor radius ($\rho$) of the UGV on the NeuroHJR performance, ablation studies are conducted with varying sensor detection ranges. The sensor radius directly influences the UGV's ability to detect obstacles and initiate evasive maneuvers, which affects both trajectory efficiency and safety margins. Each scenario maintains the same environment configuration with ten randomly placed cylindrical obstacles. Three different cases are evaluated to assess the performance of obstacle detection using sensor radii $\rho$ = 3, 5, and 7 meters, respectively. The respective trajectories are shown in Fig. \ref{fig:ablation_results}.

\begin{table}[!h]
\centering
\begin{tabular}{|c|p{3.5cm}|p{3.5cm}|}
\hline
Sensor Radius & Time taken (s) & Path Length (m)\\
\hline
3-meter & 80.6 & 103.23\\
5-meter & 77.3 & 100.48\\
7-meter & 72.2 & 95.36\\
\hline
\end{tabular}
\caption{Analysis of Trajectories with Varying Sensor Ranges}
\label{tab:sensor_radius}
\end{table}
The Table~\ref{tab:sensor_radius} shows how sensor radius affects time and path length in trajectory analysis. As the sensor range increases, there is a noticeable trend where both the time required and the path length traversed tend to decrease. This suggests a trade-off between sensor range, travelling time, and path efficiency. Based on the analysis given in this section, we conclude our study in Section \ref{sec:conclusions}.

\balance

\section{Conclusions}\label{sec:conclusions}
In this work, we introduced NeuroHJR, a novel Physics-Informed Neural Network (PINN) framework designed to approximate Hamilton-Jacobi Reachability (HJR) solutions for obstacle avoidance in cluttered environments. By embedding system dynamics and safety constraints into the loss function, NeuroHJR avoids grid-based discretization and overcomes the scalability limits of conventional HJR solvers. Our simulation results demonstrate that the proposed method achieves safety performance on par with the classical HJR approach while offering a substantial reduction of around 25\% in the travel time of the UGV, enabling real-time deployment. NeuroHJR enables the UGV to travel a shorter distance to reach the goal while avoiding obstacles. The ablation study shows how changing the sensor radius affects the overall performance of NeuroHJR. NeuroHJR represents a key step toward combining formal safety guarantees with the scalability of learning-based control. Future work will extend it to higher-dimensional systems and validate it on physical robots in complex environments.

\addtolength{\textheight}{-5cm}   


\bibliographystyle{IEEEtran}
\bibliography{ref}

\end{document}